\documentclass[letterpaper, 10 pt, conference]{ieeeconf}  

\IEEEoverridecommandlockouts                              

\usepackage{cite}
\usepackage[cmex10]{amsmath}
\usepackage{algorithmic}
\usepackage{array}
\usepackage{float}
\usepackage{graphics} 
\usepackage{epsfig} 
\usepackage{mathptmx} 
\usepackage{times} 
\usepackage{amsmath} 
\usepackage{amssymb}  
\usepackage{dblfloatfix}
\usepackage{url}
\usepackage{subcaption}
\usepackage{lipsum}
\usepackage{adjustbox,lipsum}
\usepackage{booktabs}
\usepackage{caption}
\usepackage{multirow}
\usepackage{tabularx,ragged2e,booktabs,caption}
\usepackage{textcomp}
\usepackage{hyphenat}
\usepackage{diagbox}
\usepackage{mathtools}
\usepackage{threeparttable}

\title{\LARGE \bf A Deep Learning Approach for Multi-View Engagement Estimation of Children in a Child-Robot Joint Attention task}

\author{Jack Hadfield$^{1,2}$, Georgia Chalvatzaki$^{2}$, Petros Koutras$^{1,2}$, \\Mehdi Khamassi$^{2,3}$, Costas S. Tzafestas$^{2}$ and Petros Maragos$^{1,2}$
	\thanks{This research work has been supported the EU-funded Project BabyRobot (H2020, grant agreement no. 687831).}%
	\thanks{$^{1}$Athena Research and Innovation Center, Maroussi 15125, Greece}%
	\thanks{$^{2}$School of ECE, National Technical Univ. of Athens, 15773 Athens, Greece
		{\tt\small
			jack.hadf@gmail.com,\newline \{pkoutras,ktzaf,maragos\}@cs.ntua.gr,\newline 
			gchal@mail.ntua.gr
		}  }%
	\thanks{$^{3}$Sorbonne Universit\'e, CNRS, Institute of Intelligent Systems and Robotics, Paris, France.
		{\tt\small
			mehdi.khamassi@upmc.fr
		}		}	
}

\begin{document}

	\maketitle
	\thispagestyle{empty}
	\pagestyle{empty}
	
	\begin{abstract}
		In this work we tackle the problem of child engagement estimation while children freely interact with a robot in their room. We propose a deep-based multi-view solution that takes advantage of recent developments in human pose detection. We extract the child's pose from different RGB-D cameras placed elegantly in the room, fuse the results and feed them to a deep neural network trained for classifying engagement levels. The deep network contains a recurrent layer, in order to exploit the rich temporal information contained in the pose data. The resulting method outperforms a number of baseline classifiers, and provides a promising tool for better automatic understanding of a child's attitude, interest and attention while cooperating with a robot. The goal is to integrate this model in next generation social robots as an attention monitoring tool during various CRI tasks both for Typically Developed (TD) children and children affected by autism (ASD).

	\end{abstract}

	\section{Introduction}

	As robots will become more integrated in modern societies, the cases of interacting with humans during daily life activities and tasks are increasing \cite{Goodrich07}. Human-Robot Interaction (HRI) refers to the communication between robots and humans. This communication can be verbal or non-verbal, remote or proximal. A special case of HRI is CRI \cite{Gordon2015}. Robots enter children's lives as companions, entertainers
	or even educators \cite{Kennedy,Saerbeck,Kirstein16,Davison}. Children are very adaptive, quick learners and familiarized
	with new technologies. They have unique communication
	skills, as they can easily convey or share complex
	information with little spoken language. In \cite{TKE+18,TFE+18,EKF+18} developed and evaluated systems a proposed, which employ multiple sensors and robots, for childrens' speech, gesture and action recognition during CRI scenarios. However, a major challenge is to acquire and maintain the child's engagement and attention in a CRI task \cite{Westlund}.
	
	Robots assisting children is of particular importance in modern research, especially for ASD mediated therapy towards the development of their social skills, \cite{Othman17}. Children affected by autism spectrum disorder (ASD) can benefit from interacting with robots, since such a CRI may help them overcome the impediments posed by face-to-face interaction with humans. Moreover, it is important that the robot's behaviour can adapt to the special needs of each specific child and maintain an identical behaviour for as long as needed in the intervention process \cite{Huijnen2017}.

	\begin{figure}[t]
		\centering
		\vspace{+0.2cm}
		\begin{subfigure}{.475\linewidth}
			\includegraphics[width= \linewidth]{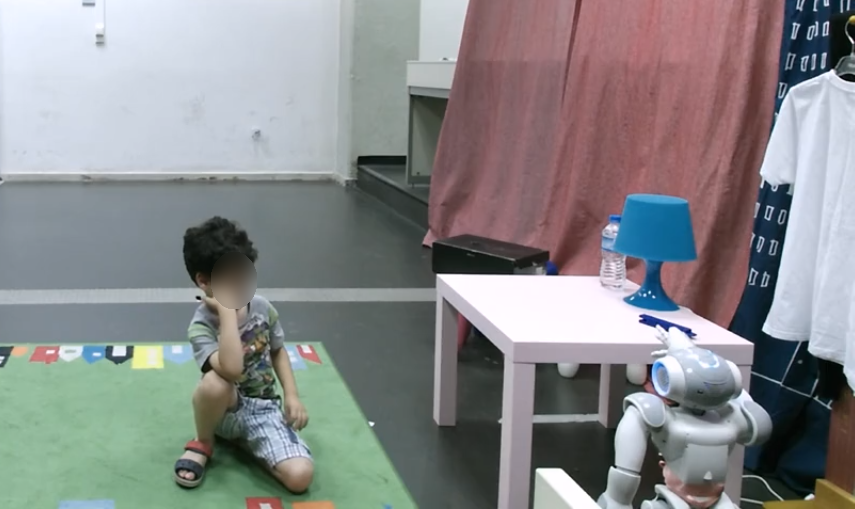}
			\caption{Class 1}
		\end{subfigure}%
		\hfill
		\begin{subfigure}{.485\linewidth}
			\includegraphics[width= \linewidth]{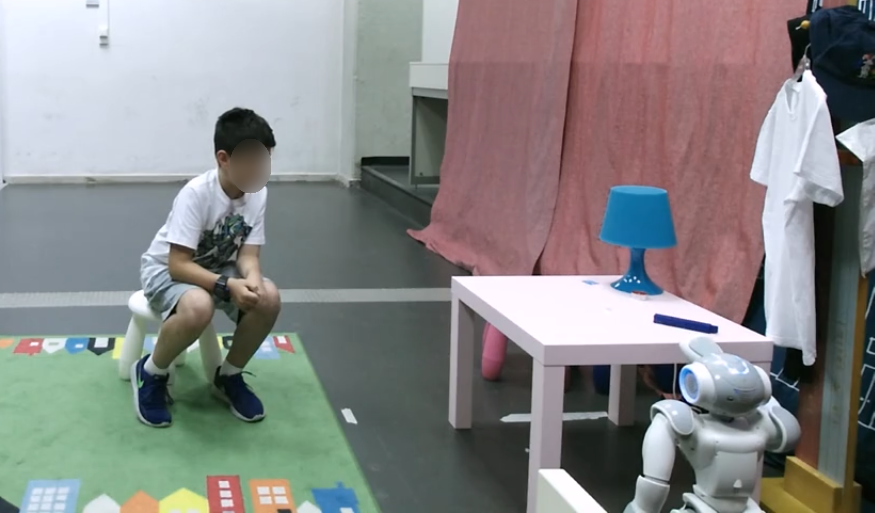}
			\caption{Class 2}
		\end{subfigure}
		\begin{subfigure}{.8\linewidth}
			\includegraphics[width=0.6 \linewidth]{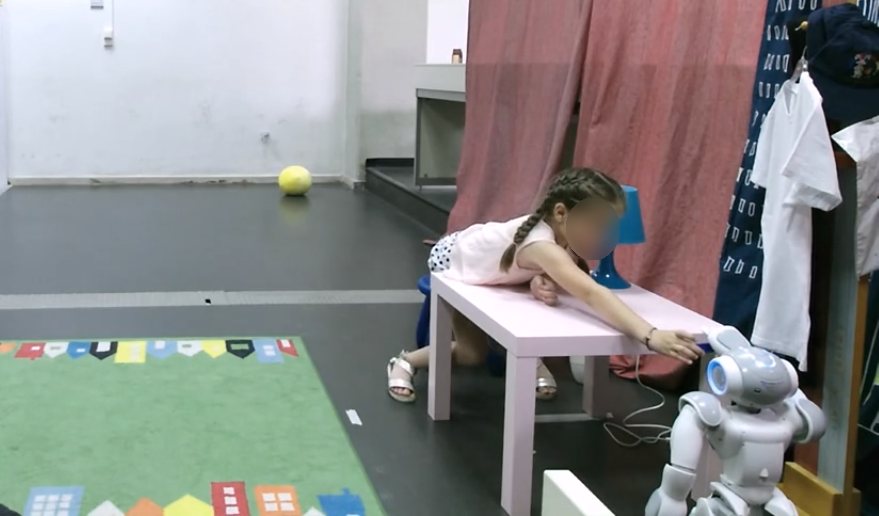}
			\centering
			\caption{Class 3}
		\end{subfigure}%
		\caption{Examples of three levels of engagement: (a) Limited attention (class 1), (b) Attention but no cooperation (class 2), (c) Active cooperation (class 3).}
		\label{fig:runtime}
	\end{figure}
	
	One key issue for social robots is the development of their ability to evaluate several aspects of interaction, such as user experience, feelings perceptions and satisfactions \cite{Anzalone2015}. Human engagement in Human-Robot Interaction (HRI) according to \cite{OBrien} ``is a category of user experience characterized by attributes of challenge, positive affect, endurability, aesthetic and sensory appeal, attention, feedback, variety/novelty, interactivity, and perceived user control''. 
	Poggi in \cite{poggi2007mind} specified more by adding that engagement is the level at which a participant attributes to the goal of being together with other participants within a social interaction and how much they continue this interaction.
	Given this rich notion of engagement, many studies have explored human-robot engagement \cite{Anzalone2015}. Lemaignan et. al explored  the level of ``with-me-ness", by measuring to what extent the human is with the robot during an interactive task, for assessing the engagement level.
	
	Research in HRI has shown a growing interest in modeling human engagement, evaluating speech and gaze \cite{Ivaldi2017}, based solely on gaze in a human-robot cooperative task \cite{Boucher}, and human pose with respect to robot from static positions \cite{Foster2017,Anzalone2015}. 
	
	Engaging children in CRI tasks is of great importance. The social characteristics that a robot should have when performing as tutors were examined in \cite{Zaga2015,Schodde2017}. Specific focus is given in estimating the engagement of children with ASD interacting with adults \cite{ami17} or robots \cite{Tapus2012}. A study analyzing the engagement of children participating in a robot-assisted therapy can be found in \cite{Rudovic17}. A method for the automatic classification of engagement with a dynamic Bayesian network using visual and acoustic cues and support vector machine classifiers is described in \cite{Feng17}. Another approach considers the facial expressions of children with ASD to evaluate their engagement \cite{Javed18}. A robot-mediated joint attention intervention system using as input the child's gaze is presented in \cite{Zheng18}. A deep learning framework for estimating the child's affective states and engagement is presented in \cite{Rudovic18}. These can then be used to optimize the CRI and monitor the
	therapy progress. In our previous work we have used reinforcement learning for adapting the robot’s behaviors and actions according to the child’s engagement, evaluated in real-time by an expert observing the child, for achieving joint attention on collaborative tasks \cite{KhamassiBailar18}.
	
	In this paper we are exploring a deep learning based approach for estimating the engagement of Children in a CRI collaborative task aiming to establish joint attention between the child and the robot. The robot tries to elicit behaviors on children while interacting with them. The robot aims to achieve joint attention with the child through an experiment that tests a primitive social skill that includes attention detection from both agents, attention manipulation from the robot-agent to the child and social coordination in terms of engaging the child-agent on a handover task, resulting in the intentional understanding of the robot-agent's intentions and ultimately a successful collaboration.

\begin{figure}[t!]
	\vspace{+0.2cm}
		\centering
		\includegraphics[width= \linewidth]{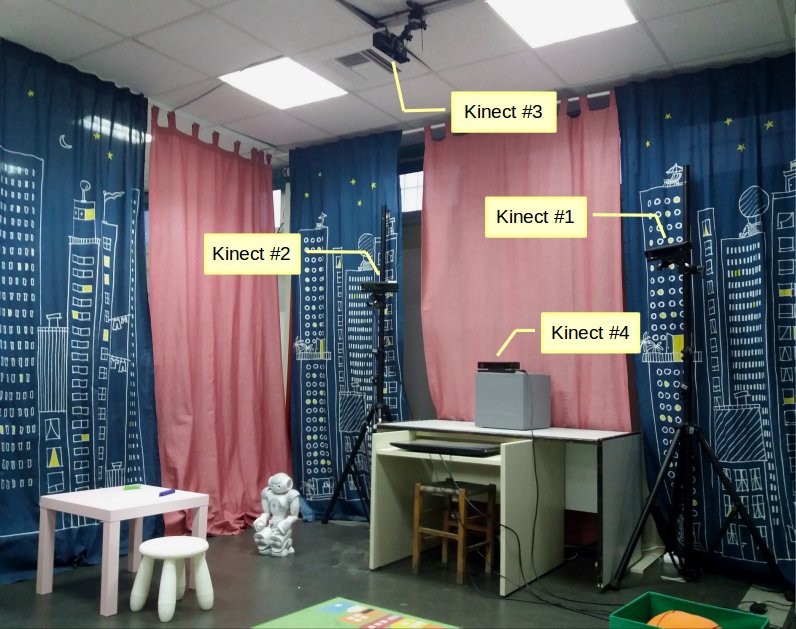}
		\caption{Setup for recording joint attention experiments.}
		\label{fig:setup}
		\vspace{-0.2cm}
	\end{figure}	
	Our method incorporates a multi-view deep-based estimation of the child's pose, when the child is inside a specially arranged room (Fig. \ref{fig:runtime}), interacts with the robot and has the ability to move freely, i.e. the child is not in a stationary position in front of the robot and the sensor. Thanks to the network of cameras, placed elegantly in the room, the child gets the feeling of being in its room and is left free to interact with the robot, without being restricted in front of a camera as in other works in literature. The multi-view fusion helps in confronting cases of body part occlusions and provides better pose estimations. An LSTM-based classifier which uses as input the child's pose is trained using as control targets observations by experts and classifies the engagement of the child to the task. 
	We experimentally validate our algorithms exploiting the RGB-D data of children who participated in the experimental scenario of the interaction task.
	
	The ultimate goal is to use this framework for estimating the engagement of children in various CRI tasks. We aim to use the engagement information in a robot reinforcement learning framework which uses the engagement monitoring estimates as a reward signal during the non-verbal social interaction, for adapting the robot's motion combinations and their level of expressivity towards maximizing the child-robot joint attention \cite{KhamassiTr18} in various collaborative tasks \cite{hadfield18} both for TD and ASD children.

\begin{figure}[t!]
		\centering
		\vspace{0.2cm}
		\hspace{-0.5cm}
		\includegraphics[width= 1.042\linewidth]{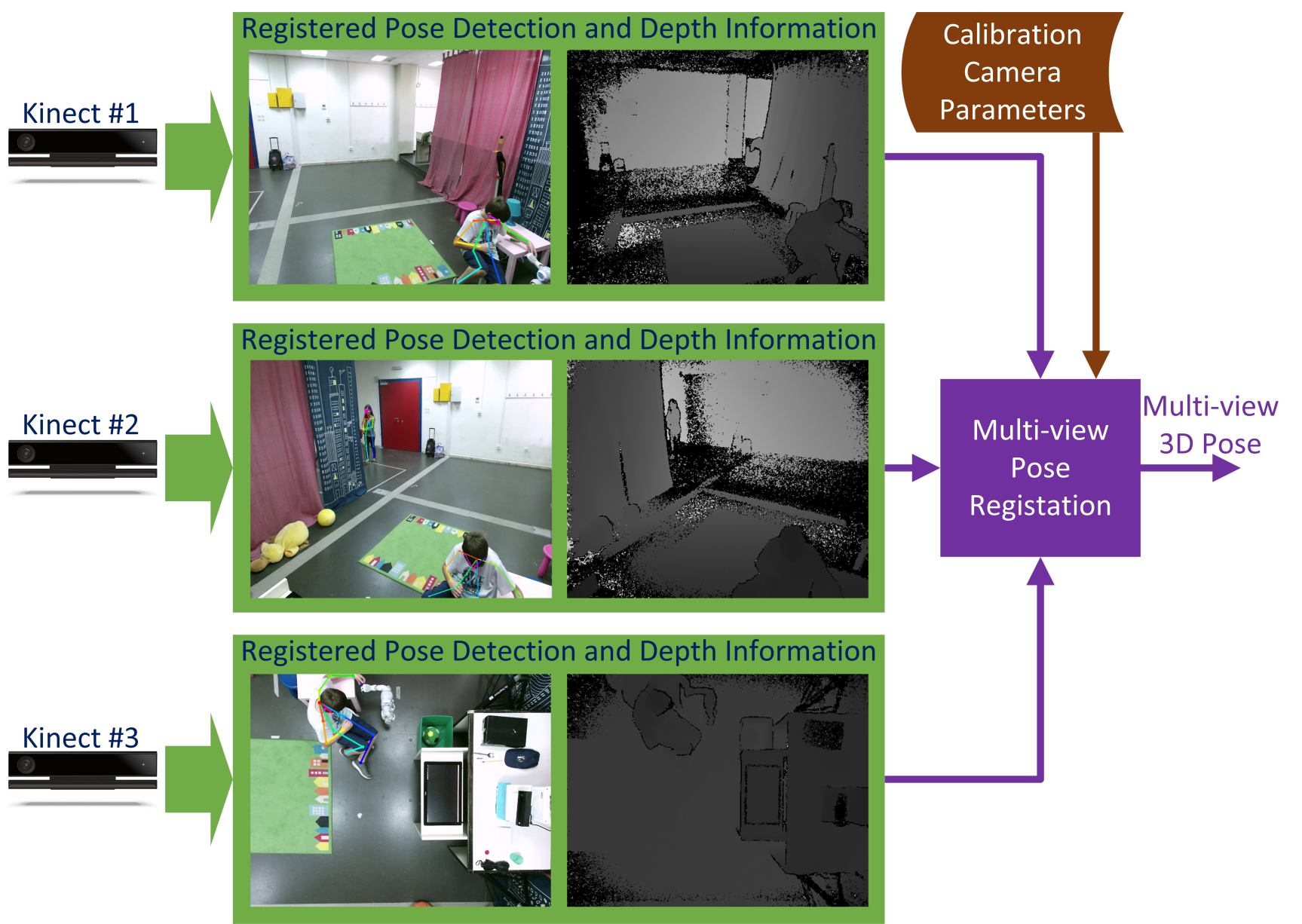}
		\vspace{0.04cm}
		\caption{Multi-view pose estimation overview.}
		\label{fig:fusion}
		\vspace{-0.2cm}
	\end{figure}

	\section{Method}
	\label{method}
	
	The main problem addressed in this work is to estimate the engagement level of children from visual cues. The problem is cast as one of multi-class classification, where each class corresponds to a different level of engagement. Specifically, we designate three distinct levels of engagement: the first (\emph{class 1}) signifies that the child is disengaged, meaning that they are paying limited or no attention to the robot; the second (\emph{class 2}) refers to a partial degree of engagement, where the child is attentive but not cooperative; the final level (\emph{class 3}) means that the child is actively cooperating with the robot to complete the handover task. The task details are described in section \ref{setup}. Of course, during the course of an interactive session, the engagement level varies. The goal is therefore to perform this classification across a number of fixed-length time segments during the session, rather than producing a single estimate for the entire interaction. In the remainder of this section we describe the proposed method to perform this classification.
	
	\subsection{Child pose estimation}

	Perhaps the most informative data for recognizing the engagement level in joint attention tasks is that of the child's pose. The problem of detecting human pose keypoints in images is a challenging one, due to occlusions and widely varying articulations and background conditions. Only recently has the problem been solved to a satisfactory degree, especially with the introduction of the Open Pose library \cite{cao2017realtime, simon2017hand, wei2016cpm} for 2D keypoint detection. Works on 3D pose estimation are fewer, with most focusing only on color images \cite{mehta2017vnect,Pavlakos_2018_CVPR}. One idea to incorporate depth information would be to estimate the 2D keypoints and take the depth values at the corresponding pixels, thus retrieving 3D coordinates. This method, however, ignores potential synergy between the two streams, and is susceptible to errors from noisy depth measurements.
		
	In \cite{zw2018}, an end-to-end 3D pose estimator from RGB-D data is proposed, which alleviates these problems and performs better than methods which operate solely on color images. The Open Pose library \cite{cao2017realtime, simon2017hand, wei2016cpm} is used to detect 2D keypoint score maps from the color image. These maps are then fed to a deep neural network, along with a voxel occupancy grid derived from the depth image. The network is trained to produce 18 keypoint estimates in the 3D space: two for each wrist, elbow, shoulder, hip, knee and ankle, one for the neck and five facial points, consisting of the ears, the eyes and the nose. We employ this system in our work, to estimate the child poses during their interaction with the robot.
	\begin{figure}[t]
		\vspace{+0.2cm}
		\centering
		\includegraphics[width= 0.9\linewidth]{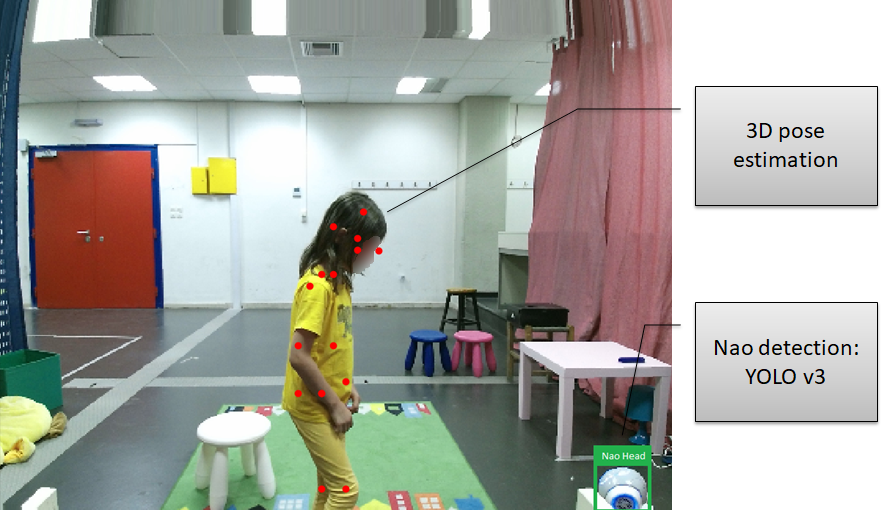}
		\vspace{0.25cm}
		\caption{The detected pose is shown along with the detected bounding box surrounding the robot's head.}
		\label{fig:detections}
		\vspace{-0.2cm}
	\end{figure}
	
	\subsection{Multi-View Fusion}
	\begin{figure}[t]
		\centering
		\vspace{+0.2cm}
		\vspace{-0.4cm}
		\includegraphics[width= 0.80\linewidth]{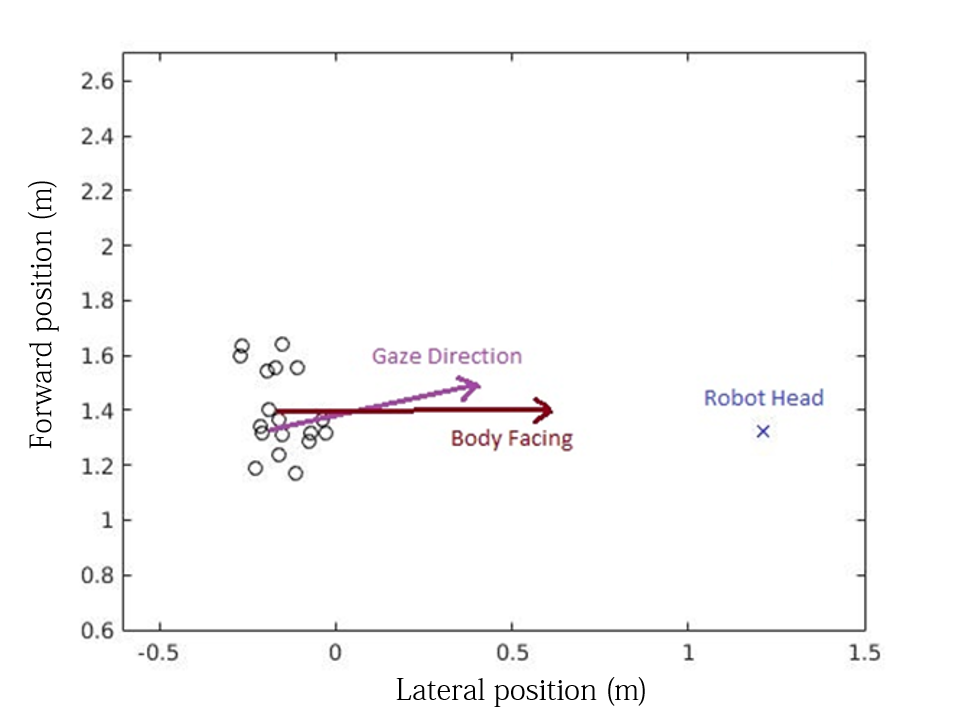}
		\vspace{-0.1cm}
		\caption{Extracted features shown from an overhead view of the room. The detected keypoints are shown as black circles.}
		\label{fig:features}
		\vspace{-0.2cm}
	\end{figure}	
	When using multiple cameras, the keypoints can be extracted for each view and fused to produce the final estimates (Fig. \ref{fig:fusion}). The first step to achieve this is to register the points of each camera reference frame to a single common frame. The registration parameters were found using the ICP algorithm \cite{besl1992method}, which provides the transformation that best fits the point cloud of one camera to that of another, given an initial transformation that we set manually. 
	
	After transforming the keypoint coordinates of all cameras to a common reference frame, the next step is to determine which keypoints are valid from each view. The pose estimation algorithm occasionally fails, either when some of the child's joints are hidden, or when the pose differs substantially from those used to train the algorithm. In such cases, the system produces noisy estimates or no detections at all for certain keypoints. Another problem is that the algorithm sometimes outputs multiple poses, when another person is in view or occasionally when the network is confused by some background artifact. To tackle such problems, we only average the points that are sufficiently close to those detected in the previous frame. If no such points exist for a certain joint, we mark the joint as missing in the current frame.
	
	Having fused the pose detections of multiple views, we interpolate the missing values using the previous estimates and then smooth the output using a simple low-pass filter. The points then undergo a final rotation, so that the coordinate axes co-align with the edges of the room.

	\subsection{Feature Extraction}
	\label{preprocessing}
	
	Naturally, we aren't interested in the child's pose in itself, but rather in relation to the task at hand. Specifically, we want to estimate the 3D keypoints in relation to the robot's position. Since the Nao robot isn't equipped with any localization sensors, we must estimate its position with respect to the world coordinates through other means. Therefore, we detect the robot in the color stream of one of the cameras, and infer its 3D position via an inverse camera projection.
	
	We fine-tuned the YOLOv3 detection network \cite{redmon2018yolov3} to detect the robot's head on a set of $100$ manually annotated images. Using this network, we then detected the robot position in all video frames. An example is shown in Fig. \ref{fig:detections}. Paired with the depth images, we converted the detections to 3D points. We then subtract the resulting values from the child pose estimates for each joint, thus making our features invariant to the position of the cameras within the experimental setup. The robot detections also contained noise and missing values, and were subjected to a similar procedure as the keypoints, ie. missing value interpolation and smoothing. We also rejected erroneous detections if they lay outside a certain expected range, based on the limitations of the robot's movement.

		\begin{figure*}[t!]
			\vspace{0.2cm}
			\begin{subfigure}{.53\linewidth}
				\centering
				\includegraphics[width=1 \linewidth]{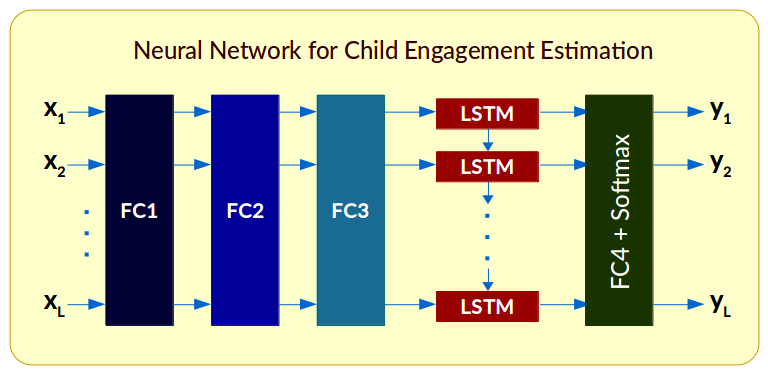}
				\caption{The neural network comprises the fully connected layers FC1, FC2, FC3, a single LSTM layer and a final fully connected layer FC4 coupled with a softmax function on the output.}
				\label{fig:network}
			\end{subfigure}%
			\hfill
			\begin{subfigure}{.43\linewidth}
				\vspace{1.92cm}
				\centering
				\begin{tabular}{ c | c | c }
					\hline
					\multicolumn{1}{c}{\bf{Layer}} & \multicolumn{1}{c}{\bf{Output}} & \multicolumn{1}{c}{\bf{Includes}} \\ \hline
					FC1 & (N,L,2C) & Dropout + ReLU\\ 
					FC2 & (N,L,2C) & Dropout + ReLU\\
					FC3 & (N,L,2C) & Dropout + ReLU\\
					LSTM & (N,L,C) & -\\
					FC4 & (N,L,K) & -\\
					\hline
				\end{tabular}
				\caption{The layer outputs sizes are shown, where N is the batch size, L is the sequence length, C is the hidden states size and K is the number of classes.}
				\label{tab:layers}
			\end{subfigure}
			\caption{(a) Architecture of neural network used to classify child engagement. (b) Network layer details. }
			\label{fig:neuralBoth}
		\end{figure*}
	We produce a number of high-level features that are expected to assist the classification process (Fig. \ref{fig:features}). These include the angle between the child's gaze and the robot, the angle between the child's body facing and the robot and the distance of the hands from the respective shoulders. The gaze direction is calculated from the detected facial keypoints, by taking the ear-to-ear vector in the 2D plane and rotating it $90\deg$. The body facing is calculated in a similar fashion from the shoulder keypoints. From the two resulting angles, we subtract the robot-to-child angle, which is calculated using the keypoint center of mass and the detected robot position. The high-level features are concatenated with the keypoint values relative to the robot, mentioned above, to form the input data to the classifier.

	\subsection{Engagement Estimation}

	A key observation worth noting is that the degree of engagement heavily depends on temporal information. One reason for this is that the child tends to display the same level of interest over the course of a few-second interval. More importantly, however, the child's movement and actions contain rich information which can be exploited. For example, if the child is constantly shifting their gaze, this is usually an indication of disengagement, whereas a steady focus signifies a higher level of interest. By choosing suitable machine learning algorithms capable of capitalizing on temporal data, we can expect a notable improvement over simply classifying each segment individually.
	
	We use a deep neural network to classify the engagement level over time, the architecture of which can be seen in Fig. \ref{fig:network}. The network consists of three fully connected (FC) layers, a Long Short-Term Memory (LSTM) network \cite{hochreiter1997long} and a final fully connected layer, with a softmax function applied to the output to produce a probability score for each class. LSTM networks are a certain type of recurrent neural network that are known to be well-suited to dealing with time-varying data. 
	
	The network is fed a sequence of $L$ inputs. We group the input features described earlier into segments of $5$ frames, over which we compute the mean and standard deviation, thus further reducing noise and avoiding training on spurious data points, which can cause over-fitting. This gives us an input vector $\textbf{x}_t$ for each segment $t \leq L$, with a dimension of
	\begin{equation}
	D=2\cdot(3\cdot18 \cdot \text{ keypoints}+4 \cdot \text{ high-level features}) = 116
	\end{equation}
	The output $\textbf{y}_t(\textbf{x}_{1 \colon t}, \textbf{W})$ is a function of the previous $t$ inputs in the sequence and the weights $\textbf{W}$ of the network, with a dimension equal to the number of classes ($K=3$). The fully connected layers produce linear combinations of their inputs, operating on each sequence point individually.
	The softmax layer ensures that the elements of $\textbf{y}_t$ are positive and sum to one. The data are fed in batches of size $N$. The output dimensions of each layer are shown in the table in Fig. \ref{fig:neuralBoth}. The variable $C$ is the size of the LSTM's hidden state. We set the output sizes of the first three fully connected layers to $2\cdot\text{C}$ after trial and error. The first three fully connected layers are each followed by a dropout layer, with a dropout probability of 0.5, and a Rectified Linear Unit (ReLU).

	\section{Experimental Analysis \& Results}
	
	\subsection{Experimental setup}
	\label{setup}

	\noindent \textbf{{1) Setup:}}
	The experimental setup can be seen in Fig. \ref{fig:setup}. As shown, $4$ Kinect devices are placed at various angles to capture multiple views of the room. Devices 1 and 2 capture the room from either side, device 3 from above and device 4 from the front.
	The Nao robot \cite{gouaillier2009mechatronic} was chosen for the task because it is capable of a wide range of motion, and its human-like features make it suitable for child-robot interaction.
	One or more bricks were placed either on a table in front of the robot, or on the floor close by. The child is free to move around the room as they please.
	
	\noindent \textbf{{2) Session Description:}}
	The experiments evolved as follows. The robot approached one of the bricks and displayed the intention of picking up the brick, without being able to actually grasp it. With a series of motions it attempted to capture the child's attention and prompt the child to hand over the brick. These motions included pointing at the brick, opening and closing its hand, alternating its gaze between the child and the brick and a combination of head turning and hand movement. If after a certain length of time the child failed to understand the robot's intent, the robot proceeded to ask the child verbally. Once the brick was successfully handed over, the robot thanked the child and in some cases looked for another brick to grasp.
	Aside from the robot's ability to communicate its intention, the success of the handover task depends on the child's visual perceptiveness, willingness to cooperate and more generally their social skills. It is evident why understanding the level of engagement that the child displays is of crucial importance, both in evaluating the child's abilities and in allowing the robot to choose more effective motions based on the child's responses.
	
	\noindent \textbf{{3) Data Collection:}}
	We recorded a total of 25 sessions. The children were aged 6-10 years old, 15 male and 10 female. The videos were then given to experts for annotation, according to the scheme described earlier. Dividing the recordings into 1 second segments, we derived 281 segments belonging to class 1, 2578 to class 2 and 745 to class 3. As we can see, the classes were significantly unbalanced: class $2$ contained around $9.2$ times more samples than class $1$ and around $3.5$ times more than class $3$.

	\subsection{Experimental Validation}

		\begin{table}[t!]
			\vspace{+0.2cm}
		\centering
		\begin{tabular}{ c | c | c | c }
			\hline
			\multicolumn{1}{c}{\bf{Method}} & \multicolumn{1}{c}{\bf{Mean F-Score}} & \multicolumn{1}{c}{\bf{Accuracy}} & \multicolumn{1}{c}{\bf{Balanced Accuracy}} \\ \hline
			Majority class & 27.90 & 71.97 & 33.33\\ 
			3FC+LSTM & \textbf{62.18} & \textbf{77.11} & \textbf{61.88}\\
			SVM & 54.79 & 68.27 & 58.61\\
			RF & 56.41 & 68.60 & 61.78\\
			\hline
		\end{tabular}
		\caption{Performance results of different algorithms on the data. Results are averaged across folds of leave-one-out cross-validation.}
		\label{tab:TDResultsAlgo}
		\vspace{-0.2cm}
	\end{table}
	We evaluate the method described above trained on the recordings of the children. Since we only have $25$ videos, rather than splitting the set into training and testing subsets, we carry out the evaluation via leave-one-out cross-validation. 
	
	\noindent \textbf{{1) Implementation:}}
	We implemented the neural network described in section \ref{method} using the PyTorch library. The network was trained from scratch, with an initial learning rate of 0.1, momentum 0.5 and weight decay $10^{-6}$. We used early stopping on a random subset of the training data, with a patience level of 10 epochs. When the training converged, we dropped the learning rate by a factor of 10. We chose a training batch size of $N=16$ and set the hidden state size to $C=560$. The sequence length $L$ was set to 30. These values were chosen after an extensive hyper-parameter search.
	
	Since LSTM networks generally require a large amount of data to train successfully and avoid over-fitting, we employ a few methods of data augmentation. Namely, we add a small amount of Gaussian noise to the mean value of each segment and randomly choose the starting point of each sequence within a range of $2$ seconds. 
	We observed a further improvement when training the fully connected layers first, and then freezing their weights, adding the LSTM module and training the remaining network. This forces the initial layers to produce informative outputs with regard to each individual segment, which the LSTM can then utilize to extract meaningful temporal information. Additionally, since we observed occasional spikes in the network's gradients, we performed gradient clipping on the LSTM layer by capping the gradient norms at the value of 0.1.

	The final classification is performed on $1$-second segments, by process of a majority vote within the segment. At a frame rate of $30$ fps, each second contains $6$ smaller segments. This seemed a logical compromise between over-sampling the data points and segmenting the temporal stream too crudely to be of use.

		\begin{table}
			\vspace{+0.2cm}
			\resizebox{\linewidth}{!}{
				\begin{tabular}{ c | c | c | c }
					\hline
					\multicolumn{1}{c}{\bf{Net Architecture}} & \multicolumn{1}{c}{\bf{Mean F-Score}} & \multicolumn{1}{c}{\bf{Accuracy}} & \multicolumn{1}{c}{\bf{Balanced Accuracy}} \\ \hline
					3 $\cdot$ FC + LSTM & \textbf{62.18} & \textbf{77.11} & \textbf{61.88}\\ 
					2 $\cdot$ FC + LSTM & 56.23 & 71.86 & 58.46\\
					3 $\cdot$ FC + 2 $\cdot$ LSTM & 54.78 & 70.60 & 56.30\\
					2 $\cdot$ FC + 2 $\cdot$ LSTM & 54.45 & 69.71 & 56.91\\
					\hline
				\end{tabular}}
				\caption{Cross-validation results for different network architectures.}
				\label{tab:TDResultsNet}
			\end{table}%
			\begin{table}
				\centering
				\begin{tabular}{ c | c | c | c }
					\hline
					\multicolumn{1}{c}{\bf{Parameters}} & \multicolumn{1}{c}{\bf{Mean F-Score}} & \multicolumn{1}{c}{\bf{Accuracy}} & \multicolumn{1}{c}{\bf{Balanced Accuracy}} \\ \hline
					N=8, L=30 & 58.75 & 74.41 & 58.40\\ 
					N=16, L=30 & \textbf{62.18} & \textbf{77.11} & \textbf{61.88}\\
					N=32, L=30 & 55.36 & 69.34 & 58.68\\ 
					N=16, L=10 & 47.21 & 61.68 & 51.58\\ 
					N=16, L=60 & 48.25 & 57.32 & 56.86\\ 
					\hline
				\end{tabular}
				\caption{Results for different hyper-parameter values.}
				\label{tab:TDResultsParam}
				
				\vspace{-0.2cm}
			\end{table}
			
	As mentioned, the classes are highly imbalanced. Though we also experimented with under-sampling and over-sampling, the best results were achieved using a weighted cross-entropy loss during training:
	\begin{equation}
	\mathcal{L} = - \frac{1}{NL} \sum_j^N \sum_t^L \textbf{w}_{c_{j,t}} \log \textbf{y}_{c_{j,t}} (\textbf{x}_{j,1 \colon t}, \textbf{W})
	\label{eq:loss}
	\end{equation}
	where $c_j$ denotes the class of the $j$-th sample in the minibatch and $\textbf{w}$ is a vector containing the weights for each class. We set $\textbf{w}=\left( 9.16, 1.00, 3.42\right) $, based on the appearance frequencies of each class in the dataset.  
	
	\noindent \textbf{{2) Evaluation Metrics:}}
	Due to the large class imbalance, the standard accuracy measure is not very informative. Therefore, we use two other measures of performance. The first is the average F-Score across all three classes, which is high only when both the precision and recall of each class is high. The second is the balanced accuracy of \cite{brodersen2010balanced}, given by:
	\begin{equation}
	\frac{1}{K} \sum_{c=1}^K \frac{TP_c}{TP_c+FP_c}
	\label{eq:baccuracy}
	\end{equation}
	where $TP_c$ and $FP_c$ denote the true and false positives respectively of class $c$, and $K=3$ is the number of classes.
	
	\noindent \textbf{{3) Results:}}
	In Table \ref{tab:TDResultsAlgo} we compare the LSTM-based network against other popular classifiers, in particular a Support Vector Machine (SVM) and a Random Forest (RF). The SVM uses an RBF kernel with a regularization weight of $C=100$ and a kernel coefficient of $\gamma=0.01$. The RF consists of 10 trees with a maximum depth of 10. The hyper-parameters of both classifiers were tuned via a grid search. As a baseline we also include the results if the majority class (class $1$) is always predicted.

	Notice that the proposed method outperforms all other classifiers, confirming our belief that exploiting temporal relations in the input data can lead to better results. The SVM and the RF both perform significantly better than simply predicting the majority class, however, meaning that even stationary pose information is partially descriptive of the engagement level.
	
	In Table \ref{tab:TDResultsNet} we evaluate some other network architectures that we also tried. We experimented with the removal of the third fully connected layer (rows 2 and 4) and with the addition of a second LSTM layer (rows 3 and 4). As shown, the chosen architecture provides notably better results across all metrics. The additional LSTM layer causes over-fitting, rather than learning any deeper information in the data. The use of dropout allows a deeper network, with the addition of the third fully connected layer boosting the performance by a large margin. 

	Finally, we provide a comparison of different hyper-parameter values in Table \ref{tab:TDResultsParam}. The optimal sequence length is small enough to allow the training set to be divided into enough sequences to avoid over-fitting, but large enough to capture long term dependencies in the data. A batch size of 16 also provides a compromise between finely sampling the training data in each iteration and avoiding local minima while training. It's worth noting that the algorithm is quite sensitive to these parameter values, possibly due to the relatively small size of the training dataset.

	As we see from the ablation study and the comparison with the other baseline methods the proposed deep network architecture can learn and track accurately the child's engagement based on its pose variation during the proposed freely interaction task. The developed system can be further improved with the presence of more annotated data and can become a useful tool for monitoring the childrens' behavior while they actively interact with robots. Note that the proposed system is designed to allow children play and interact with no motion constrains in a whole room rather than sitting in front of a robotic agent. The whole engagement module can be integrated alongside with child's speech, action or emotion recognition modules in order to create next generation social robots that can feel and understand the childrens' behavior.
	
	\section{Conclusions \& Future Work}
In this work we proposed, by taking advantage of recent progress in deep learning, an end-to-end method of child engagement estimation during child-robot collaboration without restricting their movement or requiring them to be tethered to the robot. The use of child pose data, in conjunction with an LSTM-based neural network, proved to be effective towards this goal. This is especially important considering the difficulty of the problem. Differences in child behavior and personality, a wide range of possible motions and actions and various technical challenges all contribute to this difficulty. The concept of engagement is not rigidly defined, with making the task hard even for humans. Dispite this, we achieve relatively high evaluation metrics across a dataset of 25 children.
	
	An important direction for future work will be to test the system on children affected by autism spectrum disorder (ASD). Such children exhibit social and communicative difficulties, but research has shown that they can benefit from interacting with robots. An important part of interaction between ASD children and robots would be to understand their degree of engagement, so the robots could monitor the children and adapt it's behavior to each child individually. Naturally, this imposes a further challenge, as ASD children act very differently to children in typical development.

	
	\section*{Acknowledgment}
The authors would like to thank the psychologists Asi- menia Papoulidi kai Christina Papailiou for annotating the engagement levels, and our colleagues Niki Efthymiou and Panagiotis Filntisis for helping carry out the experiments and for their technical assistance.
	
	\bibliography{Ref_Baby_ICRA19}
	\bibliographystyle{IEEEtran}

\end{document}